\documentclass[letterpaper, 10 pt, conference]{ieeeconf}  

\IEEEoverridecommandlockouts  

\usepackage{times}
\usepackage{cite}
\usepackage{multicol}
\usepackage{graphics} 
\usepackage{amsmath} 
\usepackage{amssymb}  
\usepackage{algorithm}
\usepackage{algpseudocode}
\usepackage{subcaption}
\usepackage[dvipsnames]{xcolor}
\usepackage{hyperref}
\hypersetup{
    colorlinks=true,
    linkcolor=black,    
    urlcolor=cyan,
    citecolor=black
    }

\usepackage{epsfig} 
\usepackage{times} 
\usepackage{amsmath} 
\usepackage{amssymb}  
\usepackage{multirow}
\usepackage{algpseudocode}

\usepackage{authblk}

\newcommand{\ours}{\textbf{MMCD}}

\title{\Large \bf MMCD: Multi-Modal Collaborative Decision-Making for Connected Autonomy with Knowledge Distillation}

\author{
  Rui Liu\textsuperscript{1}\thanks{\textsuperscript{1}University of Maryland, College Park, \texttt{\{ruiliu, tokekar, lin\}@umd.edu}. 
  \textsuperscript{2}North Carolina State University, \texttt{\{zwang248, pgao5\}@ncsu.edu}. 
  \textsuperscript{3}Adobe Research, \texttt{shenyu@adobe.com}.}, 
  Zikang Wang\textsuperscript{2}, 
  Peng Gao\textsuperscript{2}, 
  Yu Shen\textsuperscript{3}, 
  Pratap Tokekar\textsuperscript{1}, 
  Ming Lin\textsuperscript{1}
}

\begin{document}

\maketitle

\begin{abstract}
Autonomous systems have advanced significantly, but challenges persist in accident-prone environments where robust decision-making is crucial. A single vehicle's limited sensor range and obstructed views increase the likelihood of accidents. Multi-vehicle connected systems and multi-modal approaches, leveraging RGB images and LiDAR point clouds, have emerged as promising solutions. However, existing methods often assume the availability of all data modalities and connected vehicles during both training and testing, which is impractical due to potential sensor failures or missing connected vehicles. To address these challenges, we introduce a novel framework MMCD (Multi-Modal Collaborative Decision-making) for connected autonomy. Our framework fuses multi-modal observations from ego and collaborative vehicles to enhance decision-making under challenging conditions. To ensure robust performance when certain data modalities are unavailable during testing, we propose an approach based on cross-modal knowledge distillation with a teacher-student model structure. The teacher model is trained with multiple data modalities, while the student model is designed to operate effectively with reduced modalities. In experiments on \textit{connected autonomous driving with ground vehicles} and \textit{aerial-ground vehicles collaboration}, our method improves driving safety by up to ${\it 20.7}\%$, surpassing the best-existing baseline in detecting potential accidents and making safe driving decisions. More information can be found on our website \href{https://ruiiu.github.io/mmcd}{https://ruiiu.github.io/mmcd}.
\end{abstract}



\section{Introduction}
Autonomous technology has rapidly evolved over the past few decades, with advancements in perception \cite{chen2015deepdriving, li2020lidar, liu2024adaptive, liu2024imrl, bhaskar2024lava}, decision-making \cite{hoel2019combining, gao2024collaborative, liu2024towards}, and control systems \cite{sun2018fast, liu2023data}. However, the deployment of autonomous vehicles still face challenges, particularly in accident-prone scenarios. These scenarios demand high robustness and reliability, as any failure in decision-making could have severe consequences. A single vehicle navigating these scenarios is prone to have accidents due to occlusions and limited sensor range. One promising solution to mitigate these risks is to have multi-vehicle connected systems \cite{rahman2021multi, talebpour2016influence, ye2019evaluating}. By sharing information, vehicles can expand their field of view and reduce the chances of accidents. Another promising direction is the use of multi-modal data \cite{el2019rgb, zhuang2021perception, xiao2020multimodal, liu2025aukt}, such as RGB images and LiDAR point clouds, to enhance the perception and decision-making capabilities of autonomous systems. Recent works have combined these two paradigms to develop multi-vehicle, multi-modal systems \cite{roy2022multi, li2022v2x, piperigkos2020cooperative, liu2025caml}, leveraging both connectivity and diverse sensor data to further improve autonomous driving performance.

\begin{figure}[t]
    \vspace{6pt}
    \centering
    \includegraphics[width=0.95\linewidth]{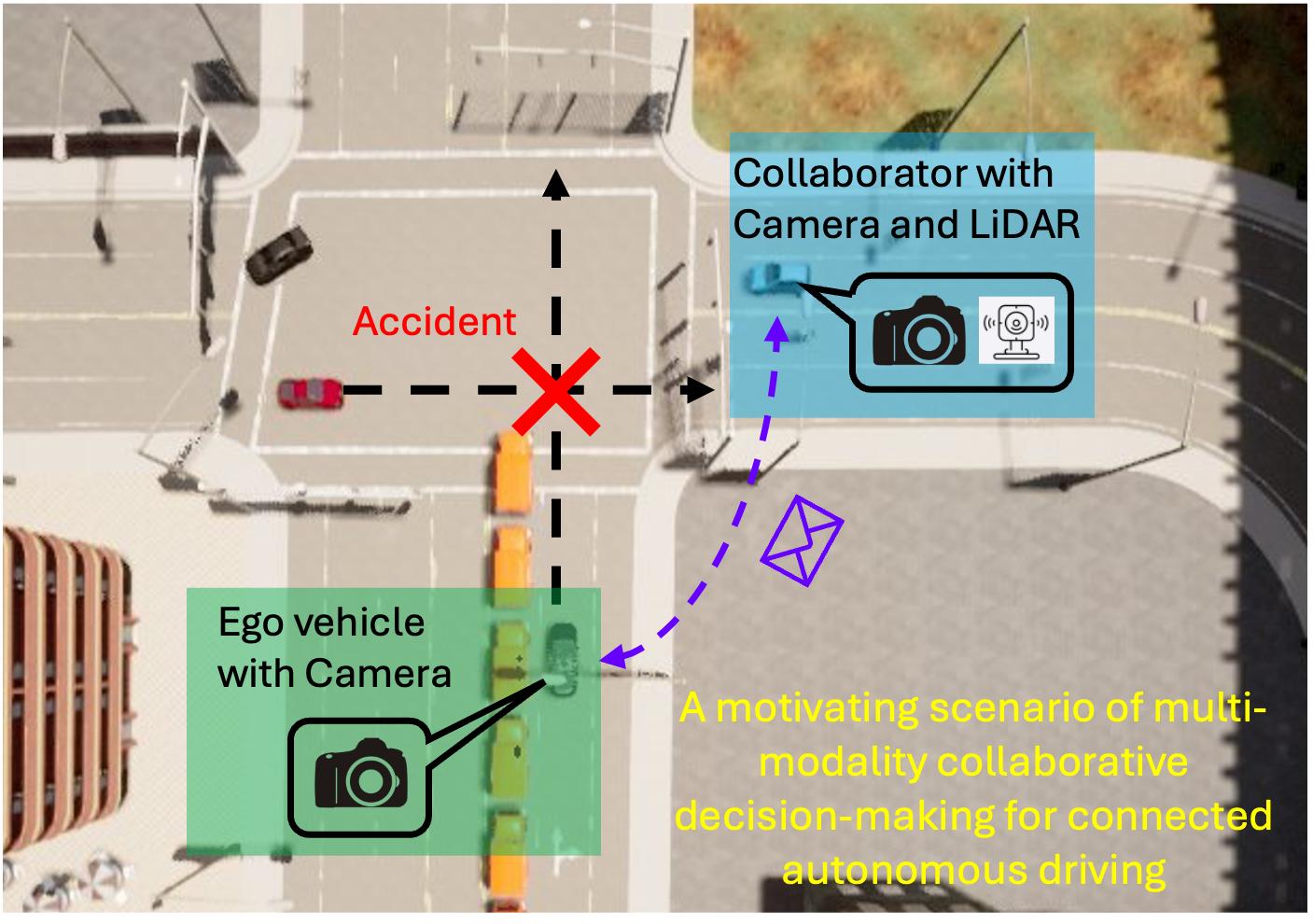}
    \caption{\textbf{A motivating scenario of multi-modal collaborative decision-making (MMCD) for connected autonomous driving}. The exchange of vital information (in purple dash line) between vehicles overcomes occlusions and sensor limitations. \ours~remains robust by leveraging available RGB data to enable the ego vehicle to take brake actions and avoid accidents, even when its LiDAR is unavailable.}
    \label{fig:mot}
    \vspace{-10pt}
\end{figure}

However, existing works often assume that the ego vehicle has consistent access to all sensors and connected vehicles during both training and testing. For example, methods utilizing both RGB and LiDAR data for training \cite{el2019rgb, zhuang2021perception, gao2018object} assume the availability of both modalities during testing. This is not always realistic; for instance, LiDAR sensors may malfunction or become unavailable during testing, leaving only RGB data accessible, or some connected vehicles may not be able to share data, as shown in Fig. \ref{fig:mot}. Additionally, cost efficiency is a crucial consideration, as LiDAR sensors are more expensive than RGB cameras. Reducing the reliance on LiDAR sensors while still achieving high performance with RGB-only models during testing presents a more cost-effective solution.

To address these challenges, we introduce a novel multi-modal collaborative decision-making framework for connected autonomy, enabling the ego vehicle to make informed decisions by leveraging shared multi-modal data from collaborative vehicles. To handle scenarios where certain data modalities are missing during testing, we propose an approach based on knowledge distillation (KD) with a teacher-student model structure. Our multi-modal framework serves as the teacher model, trained with multiple data modalities (e.g., RGB and LiDAR), while the student model operates with reduced modalities (e.g., RGB). The knowledge distillation process ensures the student model maintains robust performance even with missing modalities during test time. 


In summary, the main contributions of this paper are: 
\begin{itemize}
    \item We introduce \ours, a novel multi-modal collaborative decision-making framework for connected autonomy. Our approach fuses single or multi-modal observations provided by ego or connected vehicles in a principled way to make decisions for the ego vehicle in accident-prone scenarios. Our method improves the driving safety by up to $\bf 20.7 \%$ in experiments on connected autonomous driving with ground vehicles and aerial-ground vehicles collaboration, outperforming the best-existing baseline.
    \item We propose a cross-modal knowledge distillation-based approach for \ours. Our model is trained with multi-modal cues (e.g., LiDAR and RGB) from connected vehicles but executes using single-modality observations (e.g., RGB). This design ensures robust performance in the presence of missing modalities during testing.
\end{itemize}

\section{Related Work} \label{sec:related}
\paragraph*{Collaborative Decision Making in Autonomous Driving} Collaborative decision-making among connected vehicles enhances safety and efficiency by enabling shared sensory data and insights, especially in scenarios with limited visibility or sensor coverage. Previous works have explored frameworks such as decentralized cooperative lane-changing \cite{nie2016decentralized}, game-theoretic approaches \cite{hang2021decision}, end-to-end LiDAR-based systems \cite{cui2022coopernaut}, and spatio-temporal graph neural networks \cite{gao2024collaborative}. Communication mechanisms such as Who2com \cite{liu2020who2com}, When2com \cite{liu2020when2com}, and Where2com \cite{hu2022where2comm} have also been developed.

However, these approaches typically assume consistent access to sensors and connected vehicles during both training and testing, which is often not feasible due to potential sensor failures or data loss. Our research addresses this gap by proposing a robust collaborative decision-making framework that ensures effective performance even when some sensors are non-functional or certain data modalities are missing during testing.


\paragraph*{Multi-Modal Learning in Autonomous Driving} The fusion of RGB images and LiDAR point clouds has become essential in modern autonomous driving systems to enhance perception in complex environments. Madawi et al. \cite{el2019rgb} introduced polar-grid mapping for RGB-to-LiDAR fusion, designing early and mid-level fusion architectures. Zhuang et al. \cite{zhuang2021perception} explored RGB and LiDAR fusion for 3D semantic segmentation, while Gao et al. \cite{gao2018object} focused on object classification. Xiao et al. \cite{xiao2020multimodal} evaluated the effectiveness of combining RGB and LiDAR for end-to-end autonomous driving compared to single-modality approaches.



However, most of these methods also assume that both data modalities are consistently available during training and testing phases. Unlike previous studies, our approach addresses this limitation by implementing a knowledge distillation strategy. In this strategy, a teacher model is trained using both RGB and LiDAR data within a collaborative framework, while a student model learns to replicate the task using only RGB data. This method ensures the model remains effective even when LiDAR data is not accessible during testing.

\paragraph*{Knowledge Distillation} Knowledge distillation (KD) transfers knowledge from a complex teacher model to a simpler student model, allowing the student to perform effectively with reduced data and model complexity. KD has been widely used in natural language processing \cite{sun2019patient, xu2024survey, hahn2019self} and computer vision \cite{gou2021knowledge, beyer2022knowledge} to compress models and lower computational costs. In autonomous driving, KD has been applied mainly for perception tasks, including self-driving visual detection \cite{lan2022instance}, cooperative perception \cite{li2023mkd}, and 3D object detection \cite{cho2023itkd, sautier2022image}. However, its use in multi-modal scenarios for collaborative decision-making remains underexplored.

To address the challenge of missing data modalities during testing, we propose an approach based on KD for collaborative decision-making in connected autonomy. Notice that traditional KD distills knowledge from a larger model to a smaller one, our method distills information from a model with multiple modalities (RGB and LiDAR) into a model that uses reduced modalities (RGB only) during testing, providing a robust solution for autonomous systems.


\begin{figure*}
    \centering
    \includegraphics[width=0.95\textwidth]{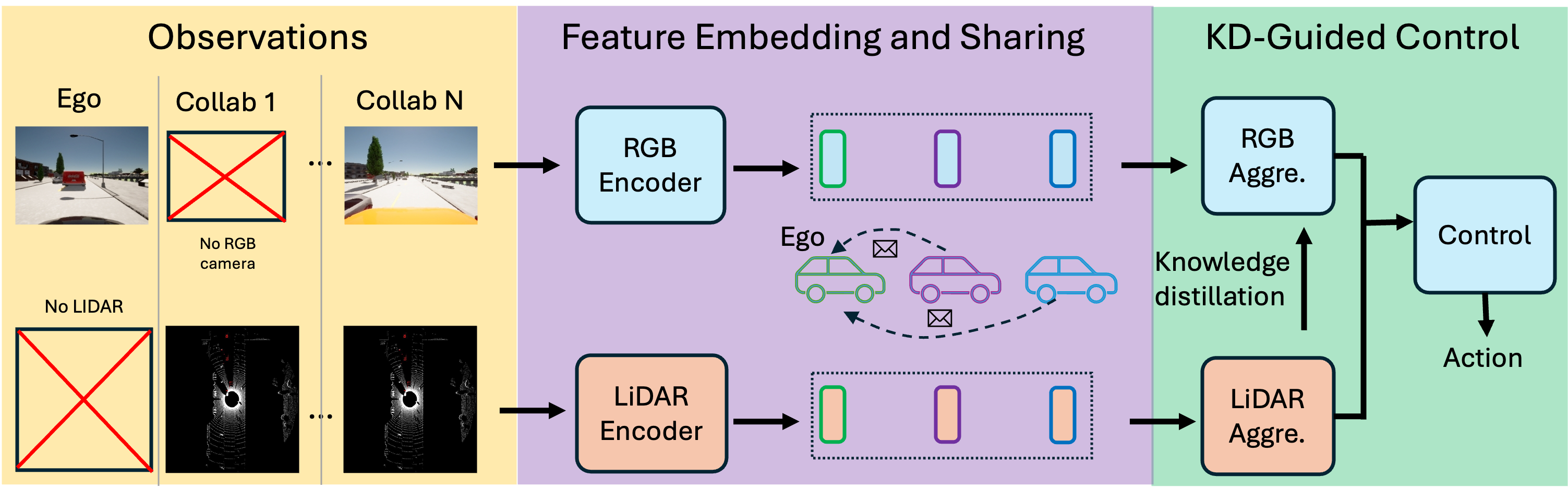}
    \caption{\textbf{Overview of the MMCD framework with knowledge distillation.} Each connected vehicle processes its own RGB data, LiDAR data, or both locally to generate feature embeddings, which are then shared to the ego vehicle. Upon receiving the shared data, the ego vehicle fuses the multi-vehicle, multi-modal features to generate its action, such as braking, particularly in accident-prone scenarios. During the training stage, LiDAR feature embeddings are used as a teacher to guide the learning of RGB feature embeddings through cross-modal knowledge distillation.}
    \label{fig:app}
    \vspace{-10pt}
\end{figure*}

\section{Approach} \label{sec:approach}
\subsection{Problem Formulation} \label{sec:prob}
In this paper, we propose a multi-modal collaborative decision-making framework for connected autonomy. Formally, we consider a setting with one ego vehicle and $N$ collaborative vehicles. The ego vehicle can receive information from the collaborative vehicles that are within a communication range $\tau$. For a time duration $T$, we represent the ego-collaborator observations as a set $\mathcal{O}_k = \{o_k^1, o_k^2, \cdots, o_k^T\}, k \in \{1, 2, \cdots, N+1\}$, where $o_k^t$ denotes the observation acquired by the $k$-th vehicle at time $t$, which can be an RGB image or a LiDAR point cloud. Given the multi-modal observations $\mathcal{O}_k$ provided by the ego and collaborative vehicles, we formulate connected autonomy as a decision-making problem, defined as $p = \pi_\theta (\mathcal{O}_k)$, where  $\pi_\theta$ denotes our policy parameterized by $\theta$.  The policy determines actions for the ego vehicle, such as braking, particularly in accident-prone scenarios. The objective is to assess whether the current situation is dangerous or not, enabling the ego vehicle to take appropriate braking or driving actions.

Given the problem formulation, we aim to address two main questions in this paper: (1) How to effectively utilize observations from connected vehicles, equipped with the same or different sensing modalities, for multi-modal collaborative decision-making? (2) How to train our model using multi-vehicle, multi-modal observations, and ensure robust performance when only single-vehicle or single-modal observations are available during testing?

\subsection{Multi-Modal Collaborative Decision Making} \label{sec:full}

To answer Question (1) of Section \ref{sec:prob} regarding multi-modal collaborative decision-making, we introduce the framework \ours, which consists of three components, including an RGB model, a LiDAR model, and a decision-making model.

\subsubsection{RGB Model} 
To process the RGB data, we develop an RGB model where each vehicle locally processes its own front-view RGB images. The model consists of two key components: an RGB encoder, which extracts visual features, and an RGB aggregator, which integrates features from connected vehicles.

\begin{figure}
    \centering
    \includegraphics[width=\linewidth]{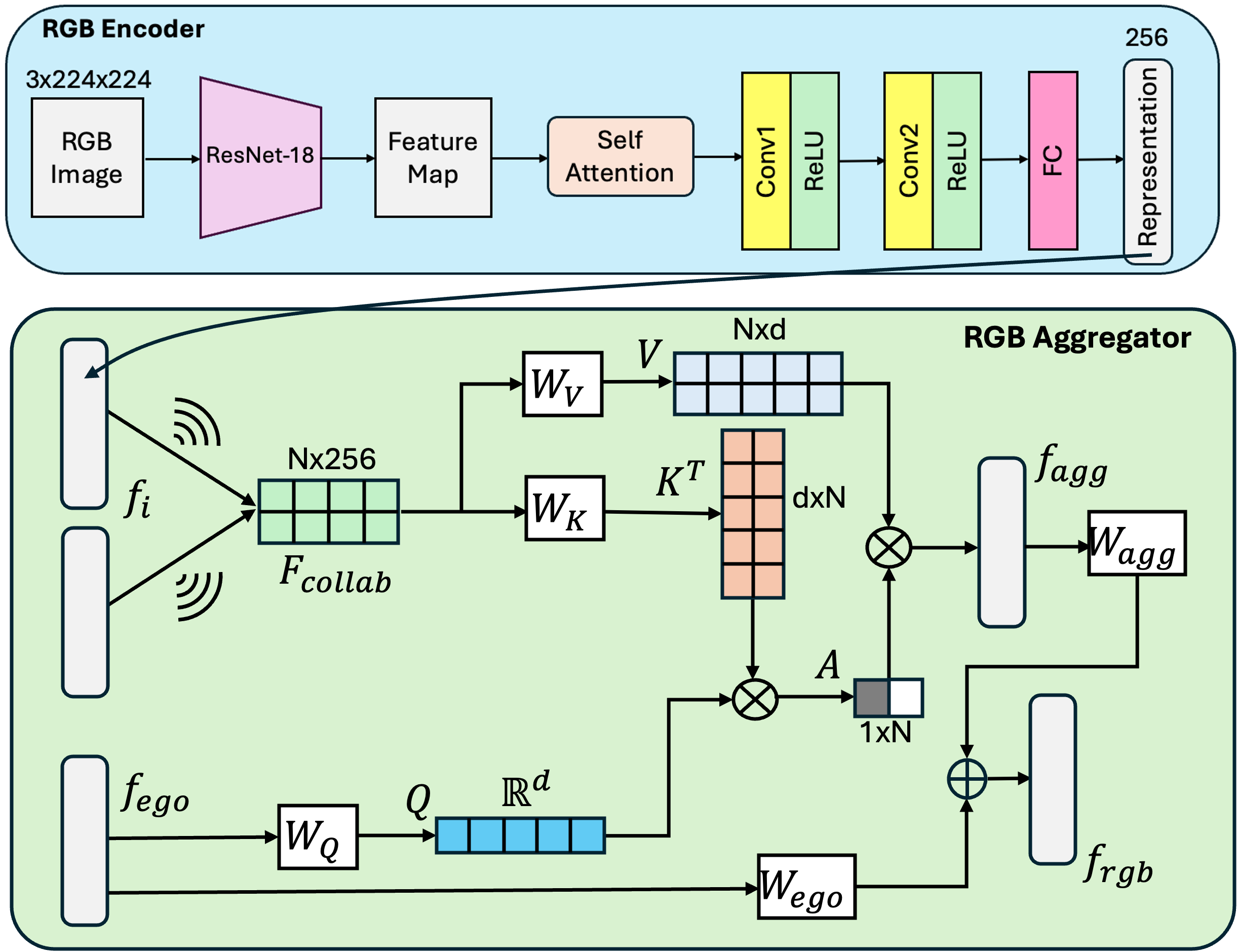}
    \caption{\textbf{RGB Model Pipeline}. The top part shows the RGB encoder and the bottom part shows the feature aggregator. Each vehicle processes its front-view RGB image using the RGB encoder to obtain feature representations. These representations from both the ego and collaborative vehicles are aggregated for the ego vehicle's decision-making. The wireless signs represent information sharing.}
    \label{fig:rgb}
    \vspace{-10pt}
\end{figure}

\paragraph{RGB Encoder}
The RGB encoder processes raw images into compact representations, which effectively captures the essential visual features needed for decision-making. Given a front RGB image $x$ of a vehicle, we first resize it to $224\times224$, then we use ResNet-18 \cite{he2016deep} as the visual backbone to extract a feature map with size $H\times W$. To allow the network to focus on the most relevant parts of the image for better decision-making, like vehicles, pedestrians, and traffic signals. We apply self-attention on the feature map to dynamically compute the importance of features at different locations. After the self-attention, we apply three convolution layers with each followed by a ReLU activation, then goes through a fully connected layer, we get a 256-dimensional feature representation, as shown in the top part of Fig. \ref{fig:rgb}.

\paragraph{RGB Aggregator}
Given the features extracted from each vehicle's front-view RGB image by the RGB encoder, we aggregate the features of all connected vehicles for collaborative decision-making, as illustrated in the bottom part of Fig. \ref{fig:rgb}. Let $\mathbf{f}^{ego} \in \mathbb{R}^{256}$ be the feature vector of the ego vehicle, $\mathbf{F}^{collab} \in \mathbb{R}^{N\times 256}$ be the matrix of feature vectors from $N$ collaborative vehicles, where each row $\mathbf{F}^{collab}_i \in \mathbb{R}^{256}$ represents a 256-dimensional feature vector from the $i$-th collaborative vehicle. To aggregate these features, we employ a cross-attention mechanism. This approach enables the ego vehicle to focus on the most important features from collaborative vehicles so that the ego vehicle can build a more informative and context-aware representation that considers the unique and dynamic contributions of each collaborative vehicle. 

We first project the input features into three spaces: query $\mathbf{Q}$, key $\mathbf{K}$, and value $\mathbf{V}$. The projections can be defined as: 
$\mathbf{Q} = \mathbf{f}^{ego} \mathbf{W^Q}$, $\mathbf{K} = \mathbf{F}^{collab} \mathbf{W^K}$, $\mathbf{V} = \mathbf{F}^{collab} \mathbf{W^V}$, 
where query vector $\mathbf{Q}$ is generated from the ego vehicle's features. Key matrix $\mathbf{K}$ and value matrix $\mathbf{V}$ are generated from the collaborative vehicles' features. $\mathbf{W^Q}, \mathbf{W^K}, \mathbf{W^V} \in \mathbb{R}^{256\times d}$ are learnable matrices. 

We then compute the attention weights, which determine the relevance of each collaborative vehicle's features to the ego vehicle's features, as defined: 
$\mathbf{A}= \text{softmax}\left(\frac{\mathbf{Q} \mathbf{K}^T}{\sqrt{d}}\right)$,
where $\mathbf{A} \in \mathbb{R}^{1\times N}$ represents the attention weights for the $N$ collaborative vehicles. We obtain the aggregated features for the ego vehicle by computing a weighted sum of the value vectors using the attention weights: 
$\mathbf{f}_{agg} = \mathbf{A} \mathbf{V}$, 
where $\mathbf{f}_{agg} \in \mathbb{R}^d$ denotes the final embedding of all RGB images provided by collaborative vehicles with the consideration of their importance for ego vehicle decision-making. 

The final RGB embedding is the combination of ego vehicle features and the aggregated features, which is defined as follows: 
$\mathbf{f}_{rgb} = \mathbf{f}_{ego}\mathbf{W_{ego}} + \mathbf{f}_{agg}\mathbf{{W_{agg}}}$, 
where $\mathbf{W_{ego}}, \mathbf{W_{agg}} \in \mathbb{R}^{d\times 256}$ are learnable parameters for transforming both the ego features and the collaborative features back to the original 256-dimensional space.

\subsubsection{LiDAR Model} For processing the LiDAR data, we utilize the COOPERNAUT \cite{cui2022coopernaut} model with two transformer blocks, which employs the Point Transformer \cite{zhao2021point} as its backbone. For each vehicle, the raw 3D point clouds are encoded into keypoints, with each keypoint associated with a representation learned by the Point Transformer blocks. The message representation of each vehicle has a size $128\times(128, 3)$. These representations from all collaborative vehicles are then merged with the ego vehicle's representations. Eventually we get the final LiDAR embedding $\mathbf{f}_{lidar}$.

\subsubsection{Decision-making Model}
If both RGB and LiDAR observations are available, we concatenate the final RGB and LiDAR embeddings for ego vehicle decision-making: $p = \mathrm{MLP}(\mathbf{f}_{rgb}||\mathbf{f}_{lidar})$, 
where $p$ is the predicted probability of braking. 
If only one modality of observation is available, either RGB or LiDAR, the decision-making model uses the available final embedding to output control actions: $p = \mathrm{MLP}(\mathbf{f}_{rgb} \ \mathrm{or} \ \mathbf{f}_{lidar})$, 
depending on the available modality.
The decision-making model enables the ego vehicle to better assess whether the current situation is dangerous and to decide accordingly between braking or continuing to drive, thereby enhancing safety in accident-prone environments.


\subsection{Cross-Modal Knowledge Distillation} \label{sec:kd}

In connected autonomy, it is common to encounter scenarios where no collaborative vehicles are present (e.g., no connected vehicles nearby) or only single-modality observations are available (e.g., surrounding connected vehicles are equipped solely with RGB cameras and lack LiDAR). As highlighted in Question (2) of Section \ref{sec:prob}, ensuring robust decision-making under such conditions is a key challenge.



To address this challenge, we propose an approach based on cross-modal knowledge distillation, as illustrated in Fig. \ref{fig:app}. We first train a teacher decision-making model $\mathcal{T}$ offline using both RGB and LiDAR data, as described in Section \ref{sec:full}, using a binary cross-entropy loss: 
$\mathcal{L}_{BCE}(y, \mathcal{T}) = -\mathbb{E}_D\big[y_i\log(p_i) + (1-y_i)\log(1-p_i) \big]$, 
where $D$ is the dataset, $y_i$ is the ground truth indicating whether the vehicle should brake, $p_i$ is the predicted probability by the teacher model.

The student model $\mathcal{S}$ is trained to mimic the behavior of the teacher model while having less modalities. For each data input, the student model receives the same RGB image that the teacher model was given. The loss for the student model is a combination of two terms: the distillation loss using KL divergence between the student output and teacher output (soft targets), and the student task loss, which is the binary cross entropy loss between the student output and the true labels (hard targets). The soft targets from the teacher enrich learning with class similarities, while hard targets ensure alignment with true labels.

The soft targets are generated by applying a temperature scaling to the logits. The scaled logits are defined as: 
$z_i = \frac{\exp{(z_i/t)}}{\exp{(z_0/t)}+\exp{(z_1/t)}}$, 
where $z_i$ is the logit for class $i$ and $t=3.0$ is the temperature. The distillation loss is defined as: 
$\mathcal{L}_{KD}(\mathcal{S}, \mathcal{T}) = -\sum z_i^{\mathcal{T}} \log(z_i^\mathcal{S})$, 
where $z_i^{\mathcal{T}}$ and $z_i^{\mathcal{S}}$ are the soft target probability from the teacher and student model, respectively. 
The overall loss for the student model is a weighted sum of the distillation loss and the binary cross-entropy loss: 
$\mathcal{L}_\mathcal{S} = (1-\alpha) \mathcal{L}_{BCE}(y, \mathcal{S}) + \alpha t^2 \mathcal{L}_{KD}(\mathcal{S}, \mathcal{T})$,
with $\alpha=0.5$ balances the two losses.


After the training of knowledge distillation process, we obtain a student model that uses only RGB data while learning from a teacher model that has access to both RGB and LiDAR data, as shown in Fig. \ref{fig:app}. This enables the student model to be effective during testing with only RGB data. Additionally, by leveraging cross-modal knowledge distillation, the student model benefits from the additional insights provided by the LiDAR data during training, learning more effectively compared to training solely with RGB data. 

\vspace{-10pt}
\section{Experiments} \label{sec:exp}
\subsection{Connected Autonomous Driving with Ground Vehicles}
To evaluate our approach, we first experiment on connected autonomous driving with ground vehicles. 

\subsubsection{Experimental Setup} \label{sec:exp_setup}
\begin{figure}[ht]
    \centering
    \includegraphics[width=\linewidth]{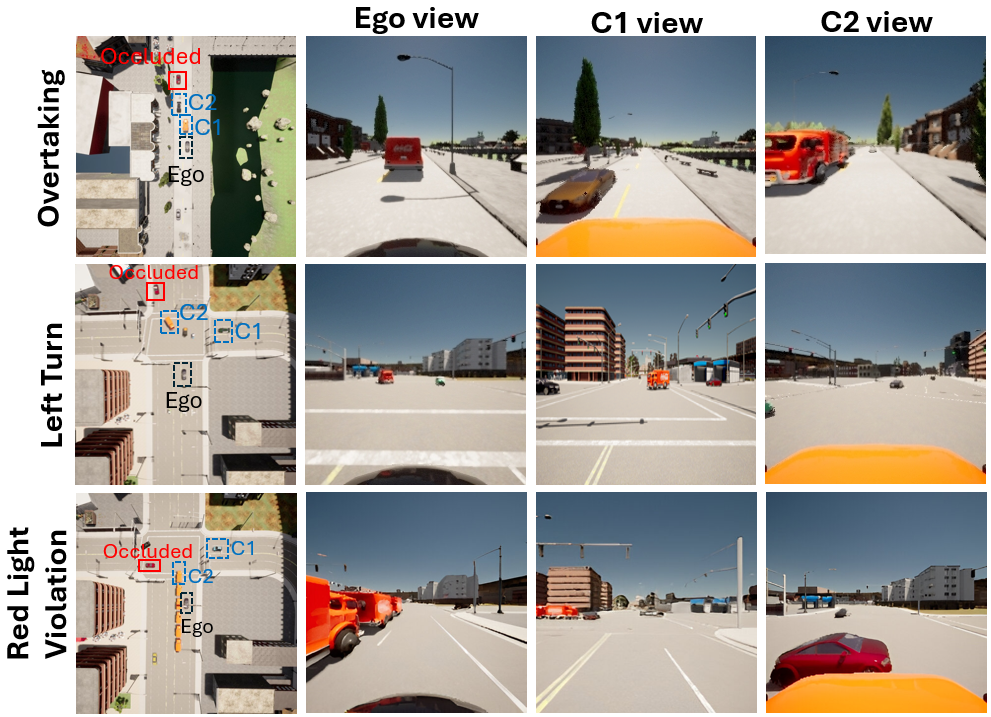}
     \caption{\textbf{Three challenging traffic scenarios prone to accidents in connected autonomous driving with ground vehicles}: overtaking, left turn, and red light violation.}
    \label{fig:scenario}
    \vspace{-10pt}
\end{figure}

Following prior studies \cite{cui2022coopernaut, gao2024collaborative}, our experiments focus on three challenging traffic scenarios with background traffic of 30 vehicles, built on CARLA \cite{dosovitskiy2017carla} integrated with AutoCast \cite{qiu2021autocast}. These scenarios are prone to accidents due to limited sensing or obstructed views, as shown in Fig. \ref{fig:scenario}, (1) \textbf{Overtaking}: A truck is blocking a sedan on a single-lane, two-way road with a dashed yellow centerline. The truck also obscures the sedan's view of oncoming traffic. The ego vehicle must decide when and how to safely change lanes to overtake the truck. (2) \textbf{Left Turn}: The ego vehicle attempts a left turn at an intersection with a yield sign. Its view is partially obstructed by another truck in the opposite left-turn lane, limiting its ability to see vehicles approaching from the opposite direction. (3) \textbf{Red Light Violation}: While crossing an intersection, the ego vehicle encounters another car that has run a red light. Due to lined-up vehicles waiting to turn left, the ego vehicle's sensors cannot detect this violator.


For each scenario, we collect 24 trails of data with the communication range $\tau=150$ m following the setup of \cite{gao2024collaborative}, with 12 used for training and 12 for testing. Each trail includes RGB images and LiDAR point clouds from up to four connected vehicles, along with the ground truth ego vehicle actions. The control module is implemented as a three-layer MLP. We train the model for 200 epochs using a batch size of 32 and the Adam optimizer \cite{kingma2014adam} with an initial learning rate of $1e{-3}$.

We consider the following four cases of data modality availability during training and testing: \textbf{Case A} (Only RGB data is available for both training and testing), \textbf{Case B} (Only LiDAR data is available for both training and testing), \textbf{Case C} (Both RGB and LiDAR data are available for both training and testing), and \textbf{Case D} (Both RGB and LiDAR data are used during training, but only RGB data is available during testing). 

Following prior work \cite{gao2024collaborative}, we employ the following metrics for evaluation: \textbf{Package Size (PS)}: Refers to the size of the shared data package between connected vehicles. It is used to evaluate the efficiency of communication in collaborative decision-making scenarios. \textbf{Accident Detection Rate (ADR)}: Defined as the ratio of detected accident-prone cases to all ground truth accident-prone cases. An accident-prone case is identified when the ego vehicle takes braking actions. This metric assesses the model's ability to detect potential accidents and make safe driving decisions. \textbf{Imitation Rate (IR)}: Represents the ratio of correctly reproduced actions to the total number of expert actions. This metric is used to evaluate the performance of the model in imitating expert driving behavior. 

\begin{table}[ht]
\centering
\caption{\textbf{Application scenarios for different approaches}. \ours~offers flexibility by supporting single-vehicle and connected-vehicle scenarios, accommodating both single-modal (RGB or LiDAR) and multi-modal (RGB and LiDAR) configurations.}
\label{tab:flex}
\resizebox{\columnwidth}{!}{%
\begin{tabular}{lcccccc}
\hline
\multicolumn{1}{c}{\multirow{2}{*}{Approach}} & \multicolumn{3}{c}{Single Vehicle}  & \multicolumn{3}{c}{Connected Vehicles}  \\ \cline{2-7} 
&\multicolumn{1}{c}{RGB} & \multicolumn{1}{l}{LiDAR} & \multicolumn{1}{c}{RGB+LiDAR} 
&\multicolumn{1}{c}{RGB} & \multicolumn{1}{l}{LiDAR} & \multicolumn{1}{c}{RGB+LiDAR} 
\\ \hline
COOPERNAUT \cite{cui2022coopernaut} & &\checkmark & & & \checkmark &  \\ 
STG \cite{gao2024collaborative} & \checkmark & & &\checkmark &  &   \\ 
AML \cite{shen2023auxiliary} & \checkmark & \checkmark & \checkmark & &  &   \\ 
\ours &  \checkmark & \checkmark & \checkmark & \checkmark & \checkmark & \checkmark \\ \hline
\end{tabular}%
}
\vspace{-10pt}
\end{table}

\begin{table*}[ht]
\centering
\caption{\textbf{Performance comparison of \ours~against multiple baselines for connected autonomous driving with ground vehicles}. PS represents communication efficiency, ADR measures driving safety, and IR evaluates expert imitation performance (metrics defined in Section \ref{sec:exp}). \ours~demonstrates superior performance across all scenarios, with improvements over the latest state-of-the-art method, STG \cite{gao2024collaborative}, highlighted in bold.}
\vspace{-5pt}
\label{tab:res}
\begin{tabular}{lccccccccccc}
\hline
& \multicolumn{4}{c}{\multirow{2}{*}{Approach}} & \multicolumn{1}{c}{\multirow{2}{*}{PS$\downarrow$}} & \multicolumn{2}{c}{Overtaking} & \multicolumn{2}{c}{Left Turn} & \multicolumn{2}{c}{Red Light Violation} \\ \cline{7-12}
\multicolumn{5}{c}{} & \multicolumn{1}{c}{} & \multicolumn{1}{c}{ADR$\uparrow$} & \multicolumn{1}{c}{IR$\uparrow$} & \multicolumn{1}{c}{ADR$\uparrow$} & \multicolumn{1}{c}{IR$\uparrow$} & \multicolumn{1}{c}{ADR$\uparrow$} & \multicolumn{1}{c}{IR$\uparrow$} \\ \hline
& \multicolumn{4}{c}{COOPERNAUT \cite{cui2022coopernaut}} & \multicolumn{1}{c}{65.5KB} & \multicolumn{1}{c}{0.8813} & \multicolumn{1}{c}{0.8544} & \multicolumn{1}{c}{0.5071} & \multicolumn{1}{c}{0.7696} & \multicolumn{1}{c}{0.5446} & \multicolumn{1}{c}{0.8323} \\
& \multicolumn{4}{c}{STG \cite{gao2024collaborative}} & \multicolumn{1}{c}{4.9KB} & \multicolumn{1}{c}{0.9265} & \multicolumn{1}{c}{0.8336} & \multicolumn{1}{c}{0.6070} & \multicolumn{1}{c}{0.7670} & \multicolumn{1}{c}{0.6451} & \multicolumn{1}{c}{0.7846} \\ 
& \multicolumn{4}{c}{AML \cite{shen2023auxiliary}} & \multicolumn{1}{c}{1.0KB} & \multicolumn{1}{c}{0.8206} & \multicolumn{1}{c}{0.8322} & \multicolumn{1}{c}{0.5000} & \multicolumn{1}{c}{0.7600} & \multicolumn{1}{c}{0.4175} & \multicolumn{1}{c}{0.7328} \\ \hline
& \multicolumn{4}{c}{Non-Collaborative} &  &  &  &  &  &  &  \\ 
& \multicolumn{2}{c}{Training} & \multicolumn{2}{c}{Testing} &  &  &  &  &  &  &  \\
& RGB & LiDAR & RGB & LiDAR &  &  &  &  &  \\ \hline
Case A &\checkmark & &\checkmark & & \multicolumn{1}{c}{1.0KB} & \multicolumn{1}{c}{0.8179} & \multicolumn{1}{c}{0.8169} & \multicolumn{1}{c}{0.4601} & \multicolumn{1}{c}{0.7268} & \multicolumn{1}{c}{0.3125} & \multicolumn{1}{c}{0.6756} \\
Case B & &\checkmark & &\checkmark & \multicolumn{1}{c}{65.5KB} & \multicolumn{1}{c}{0.8228} & \multicolumn{1}{c}{0.8134} & \multicolumn{1}{c}{0.4059} & \multicolumn{1}{c}{0.7217} & \multicolumn{1}{c}{0.4018} & \multicolumn{1}{c}{0.7351} \\ 
Case C &\checkmark &\checkmark &\checkmark &\checkmark & \multicolumn{1}{c}{66.5KB} & \multicolumn{1}{c}{0.8654} & \multicolumn{1}{c}{0.8205} & \multicolumn{1}{c}{0.5576} & \multicolumn{1}{c}{0.7483} & \multicolumn{1}{c}{0.5250} & \multicolumn{1}{c}{0.7504} \\ \hline
& \multicolumn{4}{c}{Collaborative} &  &  &  &  &  &  &  \\
& \multicolumn{2}{c}{Training} & \multicolumn{2}{c}{Testing} &  &  &  &  &  &  &  \\
& RGB & LiDAR & RGB & LiDAR &  &  &  &  &  \\ \hline
Case A &\checkmark & &\checkmark & & \multicolumn{1}{c}{1.0KB} & \multicolumn{1}{c}{0.8522} & \multicolumn{1}{c}{0.8415} & \multicolumn{1}{c}{0.5642} & \multicolumn{1}{c}{0.7804} & \multicolumn{1}{c}{0.5176} & \multicolumn{1}{c}{0.8104} \\
Case B \cite{cui2022coopernaut} & &\checkmark & &\checkmark & \multicolumn{1}{c}{65.5KB} & \multicolumn{1}{c}{0.8813} & \multicolumn{1}{c}{0.8544} & \multicolumn{1}{c}{0.5071} & \multicolumn{1}{c}{0.7696} & \multicolumn{1}{c}{0.5446} & \multicolumn{1}{c}{0.8323} \\ 
\ours &\checkmark &\checkmark &\checkmark &\checkmark & \multicolumn{1}{c}{66.5KB} & \multicolumn{1}{c}{\textbf{0.9604}} & \multicolumn{1}{c}{\textbf{0.8676}} & \multicolumn{1}{c}{\textbf{0.7329}} & \multicolumn{1}{c}{\textbf{0.8122}} & \multicolumn{1}{c}{\textbf{0.6875}} & \multicolumn{1}{c}{\textbf{0.8578}} \\
Case D &\checkmark &\checkmark &\checkmark & & \multicolumn{1}{c}{1.0KB} & \multicolumn{1}{c}{0.9288} & \multicolumn{1}{c}{0.8381} & \multicolumn{1}{c}{0.6632} & \multicolumn{1}{c}{0.7946} & \multicolumn{1}{c}{0.6607} & \multicolumn{1}{c}{0.8262} \\ \hline
& \multicolumn{4}{c}{Improvement (w.r.t. STG\cite{gao2024collaborative})} & \multicolumn{1}{c}{\textbf{4.9X}} & \multicolumn{1}{c}{$\textbf{3.7\%}$} & \multicolumn{1}{c}{$\textbf{4.1\%}$} & \multicolumn{1}{c}{$\textbf{20.7\%}$} & \multicolumn{1}{c}{$\textbf{5.9\%}$} & \multicolumn{1}{c}{$\textbf{6.6\%}$} & \multicolumn{1}{c}{$\textbf{9.3\%}$} \\ \hline
\end{tabular}
\vspace{-10pt}
\end{table*}

\subsubsection{Baselines}
We evaluate the performance of \ours~against multiple baselines, including:  
\begin{itemize}
    \item \textbf{COOPERNAUT} \cite{cui2022coopernaut}: A LiDAR-based collaborative method that utilizes LiDAR data for both training and testing. This method is functionally equivalent to our approach in collaborative settings under data modality Case B.
    \item \textbf{STG} \cite{gao2024collaborative}: A collaborative decision-making approach that leverages spatial-temporal graphs with RGBD data for both training and testing in connected autonomy.
    \item \textbf{AML} \cite{shen2023auxiliary}: A method that utilizes RGB and LiDAR data during training but relies solely on RGB data during testing. This approach operates in a single-agent context without collaboration.
    \item \textbf{Non-Collaborative Settings}: Performance evaluation across different data modalities without leveraging connected vehicle collaboration.
    \item \textbf{Collaborative Settings}: An assessment of performance across different data modalities and the robustness of our approach when certain modalities are unavailable during testing.
\end{itemize}


\subsubsection{Experimental Results}
We first outline the application scenarios for different approaches in Table \ref{tab:flex}, emphasizing the flexibility of \ours~in handling both single and multi-vehicle settings with single or multi-modal inputs. For comparison, we consider the best-performing existing methods, COOPERNAUT \cite{cui2022coopernaut}, STG \cite{gao2024collaborative} and AML \cite{shen2023auxiliary}. COOPERNAUT is limited to LiDAR data, while STG is restricted to RGB data and AML operates only in a single-vehicle setting. In contrast, \ours~supports all scenarios, accommodating both single-vehicle and connected-vehicle setups with single-modality (RGB or LiDAR) and multi-modality (RGB and LiDAR) configurations.

\paragraph{Non-Collaborative Settings}
In non-collaborative settings, the ego vehicle operates independently without any interaction or data sharing with other vehicles. The ego vehicle relies solely on its own sensor inputs to make driving decisions. We evaluate different cases based on the data modalities available during training and testing. We present the quantitative results for these cases in Table \ref{tab:res}. It is evident that Case C achieves the highest performance, utilizing multi-modal data of RGB and LiDAR data during both training and testing. AML \cite{shen2023auxiliary} shows lower performance than Case C, as it uses the student model where LiDAR data is missing during testing, only RGB is used. However, AML still outperforms Case A, indicating that incorporating auxiliary LiDAR data during training allows the model to learn more robust features. As a result, even when only RGB data is available during testing, the model performs better than when trained solely on RGB data.

\paragraph{Collaborative Settings}
We evaluate different cases of data modality and present their quantitative results in Table \ref{tab:res}. The results show a pattern similar to that observed in non-collaborative settings. \ours~achieves the best performance, as it leverages the multi-modal multi-vehicle framework with RGB and LiDAR data during both training and testing. Case D, which uses the student model with missing LiDAR data during testing, performs worse than \ours~but still better than Case A. This again demonstrates the advantage of training with auxiliary LiDAR data, where the model can generalize better with RGB-only data during testing than a model trained solely on RGB data.


It is evident that the performances of all data modality cases in collaborative settings are higher than those of non-collaborative settings, as collaborative settings provide the ego vehicle with additional information from connected vehicles, helping it to handle challenges like limited sensing or obstructed views more effectively, particularly in accident-prone scenarios. This enables the ego vehicle to make more informed and safer driving decisions.

We also compare our results with those from COOPERNAUT \cite{cui2022coopernaut} and STG \cite{gao2024collaborative}. As shown in Table \ref{tab:res}, our approach achieves a notable improvement of up to $20.7\%$ in accident detection compared to STG \cite{gao2024collaborative}.


\subsubsection{Ablation Studies}
We conduct ablation studies to analyze critical design choices and some components of our approach. We first assess the effectiveness of self-attention in the RGB encoder. We compare the performance of \ours~with and without self-attention in Collaborative Case A, where only RGB data is used for both training and testing. The results, shown in Table \ref{tab:self-att}, indicate a decrease in decision-making capability when self-attention is removed. This demonstrates that self-attention helps the RGB encoder focus on the most relevant parts of the image for better decision-making. 

Next, we compare the performance of our approach using a cross-attention mechanism versus concatenation for RGB feature aggregation in Collaborative Case A. As shown in Table \ref{tab:cross-att}, the use of cross-attention for RGB feature aggregation outperforms concatenation. This result verifies that cross-attention enables the ego vehicle to focus on the most important features from collaborative vehicles, allowing it to build a more informative and context-aware representation compared to simple concatenation.

\begin{table}[ht]
\caption{\textbf{Comparison of \ours~with and without Self-Attention in the RGB encoder}. This illustrates that incorporating self-attention enhances decision-making performance.}
\centering
\resizebox{\columnwidth}{!}{
\begin{tabular}{lcccccc}
\hline
\multirow{2}{*}{} & \multicolumn{2}{c}{Overtaking} & \multicolumn{2}{c}{Left Turn} & \multicolumn{2}{c}{Red Light Violation} \\ \cline{2-7} 
 & \multicolumn{1}{c}{ADR$\uparrow$} & \multicolumn{1}{c}{IR$\uparrow$} & \multicolumn{1}{c}{ADR$\uparrow$} & \multicolumn{1}{c}{IR$\uparrow$} & \multicolumn{1}{c}{ADR$\uparrow$} & \multicolumn{1}{c}{IR$\uparrow$} \\ \hline
With & \multicolumn{1}{c}{0.8522} &0.8415  & \multicolumn{1}{c}{0.5642} &0.7804  & \multicolumn{1}{c}{0.5176} &0.8104  \\ 
Without & \multicolumn{1}{c}{0.8319} &0.8345  & \multicolumn{1}{c}{0.5489} &0.7559  & \multicolumn{1}{c}{0.4853} &0.7778  \\ \hline
\end{tabular}
}
\label{tab:self-att}
\end{table}

\begin{table}[ht]
\caption{\textbf{Cross-Attention vs. Concatenation in \ours}. Cross-attention enables the ego vehicle to extract key features from collaborators, yielding better performance than concatenation.}
\centering
\resizebox{\columnwidth}{!}{
\begin{tabular}{lcccccc}
\hline
\multirow{2}{*}{Method} & \multicolumn{2}{c}{Overtaking} & \multicolumn{2}{c}{Left Turn} & \multicolumn{2}{c}{Red Light Violation} \\ \cline{2-7} 
 & \multicolumn{1}{c}{ADR$\uparrow$} & \multicolumn{1}{c}{IR$\uparrow$} & \multicolumn{1}{c}{ADR$\uparrow$} & \multicolumn{1}{c}{IR$\uparrow$} & \multicolumn{1}{c}{ADR$\uparrow$} & \multicolumn{1}{c}{IR$\uparrow$} \\ \hline
Cross-attention & \multicolumn{1}{c}{0.8522} &0.8415  & \multicolumn{1}{c}{0.5642} &0.7804  & \multicolumn{1}{c}{0.5176} &0.8104  \\ 
Concatenation & \multicolumn{1}{c}{0.8367} & 0.8379 & \multicolumn{1}{c}{0.5578} &0.7657   & \multicolumn{1}{c}{0.4986} & 0.7851  \\ \hline
\end{tabular}
}
\label{tab:cross-att}
\vspace{-10pt}
\end{table}

\begin{table*}[ht]
\centering
\caption{\textbf{Performance comparison of \ours~against multiple baselines for aerial-ground vehicles collaboration}. PS represents communication efficiency, ADR measures driving safety, and IR evaluates expert imitation performance (metrics defined in Section \ref{sec:exp}). \ours~shows superior performance across all evaluated scenarios and the improvement relative to the latest work, STG \cite{gao2024collaborative}, is shown in bold.}
\vspace{-5pt}
\label{tab:res2}
\begin{tabular}{lccccccccccc}
\hline
& \multicolumn{4}{c}{\multirow{2}{*}{Approach}} & \multicolumn{1}{c}{\multirow{2}{*}{PS$\downarrow$}} & \multicolumn{2}{c}{Scenario 1} & \multicolumn{2}{c}{Scenario 2} & \multicolumn{2}{c}{Scenario 3} \\ \cline{7-12}
\multicolumn{5}{c}{} & \multicolumn{1}{c}{} & \multicolumn{1}{c}{ADR$\uparrow$} & \multicolumn{1}{c}{IR$\uparrow$} & \multicolumn{1}{c}{ADR$\uparrow$} & \multicolumn{1}{c}{IR$\uparrow$} & \multicolumn{1}{c}{ADR$\uparrow$} & \multicolumn{1}{c}{IR$\uparrow$} \\ \hline
& \multicolumn{4}{c}{COOPERNAUT \cite{cui2022coopernaut}} & \multicolumn{1}{c}{65.5KB} & \multicolumn{1}{c}{0.9033} & \multicolumn{1}{c}{0.8703} & \multicolumn{1}{c}{0.8543} & \multicolumn{1}{c}{0.8617} & \multicolumn{1}{c}{0.7821} & \multicolumn{1}{c}{0.8359} \\
& \multicolumn{4}{c}{STG \cite{gao2024collaborative}} & \multicolumn{1}{c}{4.9KB} & \multicolumn{1}{c}{0.9087} & \multicolumn{1}{c}{0.8621} & \multicolumn{1}{c}{0.8410} & \multicolumn{1}{c}{0.8809} & \multicolumn{1}{c}{0.8059} & \multicolumn{1}{c}{0.8376} \\ 
& \multicolumn{4}{c}{AML \cite{shen2023auxiliary}} & \multicolumn{1}{c}{1.0KB} & \multicolumn{1}{c}{0.8333} & \multicolumn{1}{c}{0.8229} & \multicolumn{1}{c}{0.7889} & \multicolumn{1}{c}{0.7704} & \multicolumn{1}{c}{0.6953} & \multicolumn{1}{c}{0.7434} \\ \hline
& \multicolumn{4}{c}{Non-Collaborative} &  &  &  &  &  &  &  \\ 
& \multicolumn{2}{c}{Training} & \multicolumn{2}{c}{Testing} &  &  &  &  &  &  &  \\
& RGB & LiDAR & RGB & LiDAR &  &  &  &  &  \\ \hline
Case A &\checkmark & &\checkmark & & \multicolumn{1}{c}{1.0KB} & \multicolumn{1}{c}{0.8201} & \multicolumn{1}{c}{0.8188} & \multicolumn{1}{c}{0.7803} & \multicolumn{1}{c}{0.7452} & \multicolumn{1}{c}{0.6893} & \multicolumn{1}{c}{0.7052} \\
Case B & &\checkmark & &\checkmark & \multicolumn{1}{c}{65.5KB} & \multicolumn{1}{c}{0.8452} & \multicolumn{1}{c}{0.8214} & \multicolumn{1}{c}{0.7959} & \multicolumn{1}{c}{0.7543} & \multicolumn{1}{c}{0.7028} & \multicolumn{1}{c}{0.7469} \\ 
Case C &\checkmark &\checkmark &\checkmark &\checkmark & \multicolumn{1}{c}{66.5KB} & \multicolumn{1}{c}{0.8788} & \multicolumn{1}{c}{0.8421} & \multicolumn{1}{c}{0.8167} & \multicolumn{1}{c}{0.7738} & \multicolumn{1}{c}{0.7231} & \multicolumn{1}{c}{0.7732} \\ \hline
& \multicolumn{4}{c}{Collaborative} &  &  &  &  &  &  &  \\
& \multicolumn{2}{c}{Training} & \multicolumn{2}{c}{Testing} &  &  &  &  &  &  &  \\
& RGB & LiDAR & RGB & LiDAR &  &  &  &  &  \\ \hline
Case A &\checkmark & &\checkmark & & \multicolumn{1}{c}{1.0KB} & \multicolumn{1}{c}{0.8832} & \multicolumn{1}{c}{0.8561} & \multicolumn{1}{c}{0.8342} & \multicolumn{1}{c}{0.8402} & \multicolumn{1}{c}{0.7676} & \multicolumn{1}{c}{0.8143} \\
Case B \cite{cui2022coopernaut} & &\checkmark & &\checkmark & \multicolumn{1}{c}{65.5KB} & \multicolumn{1}{c}{0.9033} & \multicolumn{1}{c}{0.8703} & \multicolumn{1}{c}{0.8543} & \multicolumn{1}{c}{0.8617} & \multicolumn{1}{c}{0.7821} & \multicolumn{1}{c}{0.8359} \\ 
\ours &\checkmark &\checkmark &\checkmark &\checkmark & \multicolumn{1}{c}{66.5KB} & \multicolumn{1}{c}{\textbf{0.9658}} & \multicolumn{1}{c}{\textbf{0.9024}} & \multicolumn{1}{c}{\textbf{0.9000}} & \multicolumn{1}{c}{\textbf{0.9462}} & \multicolumn{1}{c}{\textbf{0.8375}} & \multicolumn{1}{c}{\textbf{0.8743}} \\
Case D &\checkmark &\checkmark &\checkmark & & \multicolumn{1}{c}{1.0KB} & \multicolumn{1}{c}{0.9252} & \multicolumn{1}{c}{0.8605} & \multicolumn{1}{c}{0.8632} & \multicolumn{1}{c}{0.8851} & \multicolumn{1}{c}{0.7907} & \multicolumn{1}{c}{0.8405} \\ \hline
& \multicolumn{4}{c}{Improvement (w.r.t. STG\cite{gao2024collaborative})} & \multicolumn{1}{c}{\textbf{4.9X}} & \multicolumn{1}{c}{$\textbf{6.3\%}$} & \multicolumn{1}{c}{$\textbf{4.7\%}$} & \multicolumn{1}{c}{$\textbf{7.0\%}$} & \multicolumn{1}{c}{$\textbf{7.4\%}$} & \multicolumn{1}{c}{$\textbf{3.9\%}$} & \multicolumn{1}{c}{$\textbf{4.4\%}$} \\ \hline
\end{tabular}
\vspace{-10pt}
\end{table*}

\subsection{Aerial-Ground Vehicles Collaboration}
To further evaluate our approach, we conduct experiments for connected autonomy with aerial-ground vehicles collaboration using AirSim \cite{shah2018airsim}. 

\begin{figure}[ht]
    \centering
    \includegraphics[width=0.9\linewidth]{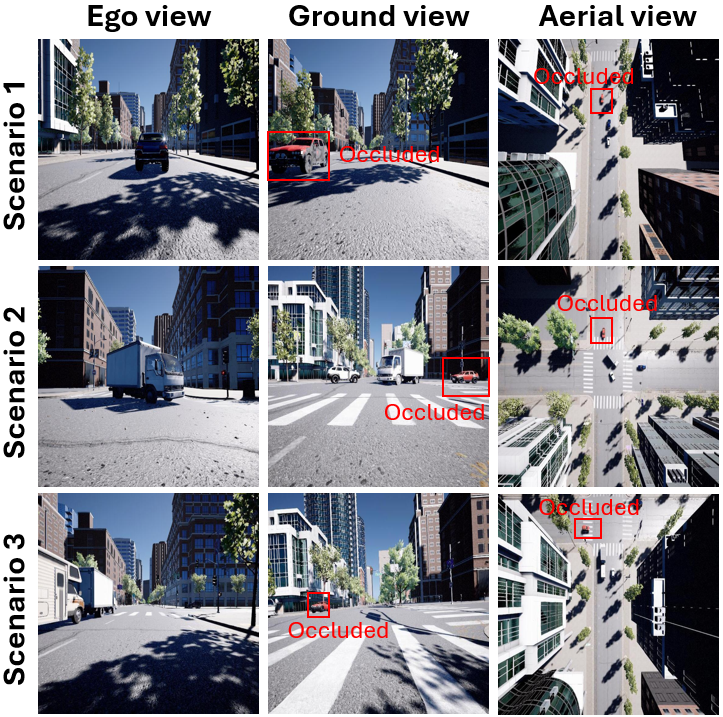}
     \caption{\textbf{Three challenging traffic scenarios prone to accidents in aerial-ground vehicle collaboration}. The views from the ego vehicle, ground collaborator, and aerial collaborator are shown, respectively. The occluded vehicle, which is not visible to the ego vehicle but can be seen by the ground and aerial collaborators, is highlighted in red.}
    \label{fig:scenario2}
    \vspace{-12pt}
\end{figure}

\subsubsection{Experimental Setup}
We consider an ego vehicle with one ground collaborator and one aerial collaborator. We focus on traffic scenarios prone to accidents due to obstructed views, similar to those studied in Section \ref{sec:exp} on connected autonomous driving with ground vehicles. These scenarios are illustrated in Fig. \ref{fig:scenario}. In Scenario 1, the ego vehicle must determine when and how to safely change lanes to overtake another vehicle. In Scenario 2, while attempting a left turn at an intersection, the ego vehicle’s view is obstructed by a truck, limiting its ability to detect oncoming traffic. In Scenario 3, as the ego vehicle crosses an intersection, another vehicle runs a red light. The presence of trucks waiting to turn left prevents the ego vehicle from detecting the violating vehicle in time. 

We use the same four data modality cases for training and testing as in Section \ref{sec:exp}. Additionally, we employ the same evaluation metrics and baselines for comparison.

\subsubsection{Experimental Results}
Similar to the findings in Section \ref{sec:exp}, the performance across all data modality cases in collaborative settings surpasses that of non-collaborative settings. This is because collaboration provides the ego vehicle with additional information from connected vehicles, allowing it to better handle obstructed views. As a result, the ego vehicle can make more informed and safer driving decisions. We also compare our approach with COOPERNAUT \cite{cui2022coopernaut}, STG \cite{gao2024collaborative}, and AML \cite{shen2023auxiliary}. As shown in Table \ref{tab:res2}, our method achieves an improvement of up to $7.0\%$ in accident detection compared to the state of the art, STG \cite{gao2024collaborative}.

Overall, the experimental results of connected autonomous driving with ground vehicles and aerial-ground vehicles collaboration demonstrate the effectiveness of \ours, and our approach can learn a policy that maintains robust performance even when there are missing data modalities during test time.  This is particularly valuable for addressing practical issues such as sensor failures. Furthermore, it offers a cost-effective solution; instead of equipping every vehicle with expensive LiDAR sensors, we can train some vehicles with both RGB and LiDAR, while deploying the model on vehicles equipped solely with RGB sensors.

\section{Conclusions} \label{sec:conclusion}
In conclusion, we introduce \ours, a novel multi-modal collaborative decision-making framework for connected autonomy. Given shared multi-modal data from connected vehicles, our framework {\em significantly improves the ego vehicle's decision-making capabilities, especially in accident-prone environments via collaborative sensing}. Experimental results demonstrate that our method improves driving safety by up to ${\bf 20.7}\%$ compared to the existing best-performing approach, highlighting its effectiveness. Additionally, we propose an approach based on cross-modal knowledge distillation, which allows the framework to maintain robust performance even when certain data modalities are unavailable during testing. This is particularly valuable for addressing practical issues of sensor failures. Furthermore, it offers a cost-effective solution for testing.

\textbf{Limitations and Future Work.} Despite the advantages, \ours~has some limitations. The current framework assumes the positions of vehicles are detected accurately for collaborative decision-making. Future work could explore state estimation strategies of vehicles under noises and uncertainties. Also, exploring the integration of other sensor modalities, such as radar or infrared, could further enhance decision-making capabilities.

\vspace*{0.5em}
\noindent
{\bf Acknowlegement:  }  This research is supported partly by ARL-UMD ArtIAMAS Cooperative Agreement, Barry Mersky \& Capital One E-Nnovate Endowed Professorships.

\bibliographystyle{ieeetr}
\bibliography{references}

\begin{thebibliography}{10}

\bibitem{chen2015deepdriving}
C.~Chen, A.~Seff, A.~Kornhauser, and J.~Xiao, ``Deepdriving: Learning affordance for direct perception in autonomous driving,'' in {\em Proceedings of the IEEE international conference on computer vision}, pp.~2722--2730, 2015.

\bibitem{li2020lidar}
Y.~Li and J.~Ibanez-Guzman, ``Lidar for autonomous driving: The principles, challenges, and trends for automotive lidar and perception systems,'' {\em IEEE Signal Processing Magazine}, vol.~37, no.~4, pp.~50--61, 2020.

\bibitem{liu2024adaptive}
R.~Liu, A.~Bhaskar, and P.~Tokekar, ``Adaptive visual imitation learning for robotic assisted feeding across varied bowl configurations and food types,'' {\em arXiv preprint arXiv:2403.12891}, 2024.

\bibitem{liu2024imrl}
R.~Liu, Z.~Mahammad, A.~Bhaskar, and P.~Tokekar, ``Imrl: Integrating visual, physical, temporal, and geometric representations for enhanced food acquisition,'' {\em arXiv preprint arXiv:2409.12092}, 2024.

\bibitem{bhaskar2024lava}
A.~Bhaskar, R.~Liu, V.~D. Sharma, G.~Shi, and P.~Tokekar, ``Lava: Long-horizon visual action based food acquisition,'' {\em arXiv preprint arXiv:2403.12876}, 2024.

\bibitem{hoel2019combining}
C.-J. Hoel, K.~Driggs-Campbell, K.~Wolff, L.~Laine, and M.~J. Kochenderfer, ``Combining planning and deep reinforcement learning in tactical decision making for autonomous driving,'' {\em IEEE transactions on intelligent vehicles}, vol.~5, no.~2, pp.~294--305, 2019.

\bibitem{gao2024collaborative}
P.~Gao, Y.~Shen, and M.~C. Lin, ``Collaborative decision-making using spatiotemporal graphs in connected autonomy,'' in {\em 2024 IEEE International Conference on Robotics and Automation (ICRA)}, pp.~4983--4989, IEEE, 2024.

\bibitem{liu2024towards}
R.~Liu, A.~Gupta, E.~Noorani, and P.~Tokekar, ``Towards efficient risk-sensitive policy gradient: An iteration complexity analysis,'' {\em arXiv preprint arXiv:2403.08955}, 2024.

\bibitem{sun2018fast}
L.~Sun, C.~Peng, W.~Zhan, and M.~Tomizuka, ``A fast integrated planning and control framework for autonomous driving via imitation learning,'' in {\em Dynamic Systems and Control Conference}, vol.~51913, p.~V003T37A012, American Society of Mechanical Engineers, 2018.

\bibitem{liu2023data}
R.~Liu, G.~Shi, and P.~Tokekar, ``Data-driven distributionally robust optimal control with state-dependent noise,'' in {\em 2023 IEEE/RSJ International Conference on Intelligent Robots and Systems (IROS)}, pp.~9986--9991, IEEE, 2023.

\bibitem{rahman2021multi}
M.~H. Rahman, M.~Abdel-Aty, and Y.~Wu, ``A multi-vehicle communication system to assess the safety and mobility of connected and automated vehicles,'' {\em Transportation research part C: emerging technologies}, vol.~124, p.~102887, 2021.

\bibitem{talebpour2016influence}
A.~Talebpour and H.~S. Mahmassani, ``Influence of connected and autonomous vehicles on traffic flow stability and throughput,'' {\em Transportation research part C: emerging technologies}, vol.~71, pp.~143--163, 2016.

\bibitem{ye2019evaluating}
L.~Ye and T.~Yamamoto, ``Evaluating the impact of connected and autonomous vehicles on traffic safety,'' {\em Physica A: Statistical Mechanics and its Applications}, vol.~526, p.~121009, 2019.

\bibitem{el2019rgb}
K.~El~Madawi, H.~Rashed, A.~El~Sallab, O.~Nasr, H.~Kamel, and S.~Yogamani, ``Rgb and lidar fusion based 3d semantic segmentation for autonomous driving,'' in {\em 2019 IEEE Intelligent Transportation Systems Conference (ITSC)}, pp.~7--12, IEEE, 2019.

\bibitem{zhuang2021perception}
Z.~Zhuang, R.~Li, K.~Jia, Q.~Wang, Y.~Li, and M.~Tan, ``Perception-aware multi-sensor fusion for 3d lidar semantic segmentation,'' in {\em Proceedings of the IEEE/CVF International Conference on Computer Vision}, pp.~16280--16290, 2021.

\bibitem{xiao2020multimodal}
Y.~Xiao, F.~Codevilla, A.~Gurram, O.~Urfalioglu, and A.~M. L{\'o}pez, ``Multimodal end-to-end autonomous driving,'' {\em IEEE Transactions on Intelligent Transportation Systems}, vol.~23, no.~1, pp.~537--547, 2020.

\bibitem{liu2025aukt}
R.~Liu, P.~Gao, Y.~Shen, M.~Lin, and P.~Tokekar, ``Aukt: Adaptive uncertainty-guided knowledge transfer with conformal prediction,'' {\em arXiv preprint arXiv:2502.16736}, 2025.

\bibitem{roy2022multi}
D.~Roy, Y.~Li, T.~Jian, P.~Tian, K.~Chowdhury, and S.~Ioannidis, ``Multi-modality sensing and data fusion for multi-vehicle detection,'' {\em IEEE Transactions on Multimedia}, vol.~25, pp.~2280--2295, 2022.

\bibitem{li2022v2x}
Y.~Li, D.~Ma, Z.~An, Z.~Wang, Y.~Zhong, S.~Chen, and C.~Feng, ``V2x-sim: Multi-agent collaborative perception dataset and benchmark for autonomous driving,'' {\em IEEE Robotics and Automation Letters}, vol.~7, no.~4, pp.~10914--10921, 2022.

\bibitem{piperigkos2020cooperative}
N.~Piperigkos, A.~S. Lalos, K.~Berberidis, and C.~Anagnostopoulos, ``Cooperative multi-modal localization in connected and autonomous vehicles,'' in {\em 2020 IEEE 3rd Connected and Automated Vehicles Symposium (CAVS)}, pp.~1--5, IEEE, 2020.

\bibitem{liu2025caml}
R.~Liu, Y.~Shen, P.~Gao, P.~Tokekar, and M.~Lin, ``Caml: Collaborative auxiliary modality learning for multi-agent systems,'' {\em arXiv preprint arXiv:2502.17821}, 2025.

\bibitem{gao2018object}
H.~Gao, B.~Cheng, J.~Wang, K.~Li, J.~Zhao, and D.~Li, ``Object classification using cnn-based fusion of vision and lidar in autonomous vehicle environment,'' {\em IEEE Transactions on Industrial Informatics}, vol.~14, no.~9, pp.~4224--4231, 2018.

\bibitem{nie2016decentralized}
J.~Nie, J.~Zhang, W.~Ding, X.~Wan, X.~Chen, and B.~Ran, ``Decentralized cooperative lane-changing decision-making for connected autonomous vehicles,'' {\em IEEE access}, vol.~4, pp.~9413--9420, 2016.

\bibitem{hang2021decision}
P.~Hang, C.~Huang, Z.~Hu, Y.~Xing, and C.~Lv, ``Decision making of connected automated vehicles at an unsignalized roundabout considering personalized driving behaviours,'' {\em IEEE Transactions on Vehicular Technology}, vol.~70, no.~5, pp.~4051--4064, 2021.

\bibitem{cui2022coopernaut}
J.~Cui, H.~Qiu, D.~Chen, P.~Stone, and Y.~Zhu, ``Coopernaut: End-to-end driving with cooperative perception for networked vehicles,'' in {\em Proceedings of the IEEE/CVF Conference on Computer Vision and Pattern Recognition}, pp.~17252--17262, 2022.

\bibitem{liu2020who2com}
Y.-C. Liu, J.~Tian, C.-Y. Ma, N.~Glaser, C.-W. Kuo, and Z.~Kira, ``Who2com: Collaborative perception via learnable handshake communication,'' in {\em 2020 IEEE International Conference on Robotics and Automation (ICRA)}, pp.~6876--6883, IEEE, 2020.

\bibitem{liu2020when2com}
Y.-C. Liu, J.~Tian, N.~Glaser, and Z.~Kira, ``When2com: Multi-agent perception via communication graph grouping,'' in {\em Proceedings of the IEEE/CVF Conference on computer vision and pattern recognition}, pp.~4106--4115, 2020.

\bibitem{hu2022where2comm}
Y.~Hu, S.~Fang, Z.~Lei, Y.~Zhong, and S.~Chen, ``Where2comm: Communication-efficient collaborative perception via spatial confidence maps,'' {\em Advances in neural information processing systems}, vol.~35, pp.~4874--4886, 2022.

\bibitem{sun2019patient}
S.~Sun, Y.~Cheng, Z.~Gan, and J.~Liu, ``Patient knowledge distillation for bert model compression,'' {\em arXiv preprint arXiv:1908.09355}, 2019.

\bibitem{xu2024survey}
X.~Xu, M.~Li, C.~Tao, T.~Shen, R.~Cheng, J.~Li, C.~Xu, D.~Tao, and T.~Zhou, ``A survey on knowledge distillation of large language models,'' {\em arXiv preprint arXiv:2402.13116}, 2024.

\bibitem{hahn2019self}
S.~Hahn and H.~Choi, ``Self-knowledge distillation in natural language processing,'' {\em arXiv preprint arXiv:1908.01851}, 2019.

\bibitem{gou2021knowledge}
J.~Gou, B.~Yu, S.~J. Maybank, and D.~Tao, ``Knowledge distillation: A survey,'' {\em International Journal of Computer Vision}, vol.~129, no.~6, pp.~1789--1819, 2021.

\bibitem{beyer2022knowledge}
L.~Beyer, X.~Zhai, A.~Royer, L.~Markeeva, R.~Anil, and A.~Kolesnikov, ``Knowledge distillation: A good teacher is patient and consistent,'' in {\em Proceedings of the IEEE/CVF conference on computer vision and pattern recognition}, pp.~10925--10934, 2022.

\bibitem{lan2022instance}
Q.~Lan and Q.~Tian, ``Instance, scale, and teacher adaptive knowledge distillation for visual detection in autonomous driving,'' {\em IEEE Transactions on Intelligent Vehicles}, vol.~8, no.~3, pp.~2358--2370, 2022.

\bibitem{li2023mkd}
Z.~Li, H.~Liang, H.~Wang, M.~Zhao, J.~Wang, and X.~Zheng, ``Mkd-cooper: Cooperative 3d object detection for autonomous driving via multi-teacher knowledge distillation,'' {\em IEEE Transactions on Intelligent Vehicles}, 2023.

\bibitem{cho2023itkd}
H.~Cho, J.~Choi, G.~Baek, and W.~Hwang, ``itkd: Interchange transfer-based knowledge distillation for 3d object detection,'' in {\em Proceedings of the IEEE/CVF Conference on Computer Vision and Pattern Recognition}, pp.~13540--13549, 2023.

\bibitem{sautier2022image}
C.~Sautier, G.~Puy, S.~Gidaris, A.~Boulch, A.~Bursuc, and R.~Marlet, ``Image-to-lidar self-supervised distillation for autonomous driving data,'' in {\em Proceedings of the IEEE/CVF Conference on Computer Vision and Pattern Recognition}, pp.~9891--9901, 2022.

\bibitem{he2016deep}
K.~He, X.~Zhang, S.~Ren, and J.~Sun, ``Deep residual learning for image recognition,'' in {\em Proceedings of the IEEE conference on computer vision and pattern recognition}, pp.~770--778, 2016.

\bibitem{zhao2021point}
H.~Zhao, L.~Jiang, J.~Jia, P.~H. Torr, and V.~Koltun, ``Point transformer,'' in {\em Proceedings of the IEEE/CVF international conference on computer vision}, pp.~16259--16268, 2021.

\bibitem{dosovitskiy2017carla}
A.~Dosovitskiy, G.~Ros, F.~Codevilla, A.~Lopez, and V.~Koltun, ``Carla: An open urban driving simulator,'' in {\em Conference on robot learning}, pp.~1--16, PMLR, 2017.

\bibitem{qiu2021autocast}
H.~Qiu, P.~Huang, N.~Asavisanu, X.~Liu, K.~Psounis, and R.~Govindan, ``Autocast: Scalable infrastructure-less cooperative perception for distributed collaborative driving,'' {\em arXiv preprint arXiv:2112.14947}, 2021.

\bibitem{kingma2014adam}
D.~P. Kingma, ``Adam: A method for stochastic optimization,'' {\em arXiv preprint arXiv:1412.6980}, 2014.

\bibitem{shen2023auxiliary}
Y.~Shen, X.~Wang, P.~Gao, and M.~Lin, ``Auxiliary modality learning with generalized curriculum distillation,'' in {\em International Conference on Machine Learning}, pp.~31057--31076, PMLR, 2023.

\bibitem{shah2018airsim}
S.~Shah, D.~Dey, C.~Lovett, and A.~Kapoor, ``Airsim: High-fidelity visual and physical simulation for autonomous vehicles,'' in {\em Field and Service Robotics: Results of the 11th International Conference}, pp.~621--635, Springer, 2018.

\end{thebibliography}
\end{document}